\documentclass{article}

\usepackage{microtype}
\usepackage{graphicx}
\usepackage{subfigure}
\usepackage{booktabs} % for professional tables

\usepackage[square,sort&compress,comma,numbers]{natbib}

\usepackage[utf8]{inputenc} % allow utf-8 input
\usepackage[T1]{fontenc}    % use 8-bit T1 fonts
\usepackage{url}            % simple URL typesetting
\usepackage{booktabs}       % professional-quality tables
\usepackage{amsfonts}       % blackboard math symbols
\usepackage{nicefrac}       % compact symbols for 1/2, etc.
\usepackage{microtype}      % microtypography
\usepackage{soul}
\usepackage[utf8]{inputenc}
\usepackage{graphicx}
\usepackage{amsmath}
\usepackage{booktabs}
\usepackage{outlines}
\usepackage{subfigure}
\usepackage{adjustbox}

\usepackage[colorlinks=true,
citecolor=blue,
filecolor=black,
linkcolor=blue,
urlcolor=blue]{hyperref}

\usepackage{microtype}
\usepackage{graphicx}
\usepackage{booktabs} % for professional tables

\usepackage{amsmath,amsfonts,bm}

\usepackage{xcolor}
\usepackage{colortbl}

\def\eqref#1{equation~\ref{#1}}
\def\ceil#1{\lceil #1 \rceil}
\def\floor#1{\lfloor #1 \rfloor}
\DeclareMathAlphabet{\mathsfit}{\encodingdefault}{\sfdefault}{m}{sl}
\SetMathAlphabet{\mathsfit}{bold}{\encodingdefault}{\sfdefault}{bx}{n}

\usepackage{multirow}
\usepackage{paralist}

\usepackage{booktabs} % for professional tables
\usepackage{multicol}
\usepackage{diagbox}

\usepackage[ruled,noend,algo2e]{algorithm2e}

\SetCommentSty{mycommfont}

\usepackage{here}

\usepackage{amsmath,amssymb,amsfonts,amsbsy,amsfonts,latexsym}
\usepackage{multirow}
\usepackage{makecell}
\usepackage[labelfont=bf,textfont=it,belowskip=0pt,aboveskip=5pt,tableposition=top]{caption}
\usepackage{xcolor}
\usepackage{colortbl}

\definecolor{colorA}{RGB}{189,201,225}
\definecolor{colorB}{RGB}{103,169,207}
\definecolor{colorC}{RGB}{ 28,144,153}
\definecolor{colorD}{RGB}{  1,108, 89}

\newcolumntype{R}{>{\columncolor{gray!40}}r}
\newcolumntype{L}{>{\columncolor{gray!40}}l}
\newcolumntype{C}{>{\columncolor{gray!40}}c}

\usepackage{tabularx,colortbl,xcolor}
\usepackage{multirow}
\usepackage[normalem]{ulem}
\useunder{\uline}{\ul}{}

\usepackage{enumitem}

\usepackage{xparse}% http://ctan.org/pkg/xparse

\DeclareGraphicsExtensions{.pdf,.png}

\SetKwInput{KwInput}{Input}
\SetKwInput{KwRequire}{Require}

\usepackage{longtable}
\usepackage{pgfplots}

\usepackage{lipsum}

\usepackage{amssymb}% http://ctan.org/pkg/amssymb
\usepackage{pifont}% http://ctan.org/pkg/pifont
\newcommand{\cmark}{\textcolor{teal}{\ding{51}}}%
\newcommand{\xmark}{\textcolor{red}{\ding{55}}}%
\usepackage{adjustbox}

\NewDocumentCommand{\var}{O{s} m O{}}{%
  \ensuremath{#1_{#2}^{#3}}% add \vphantom{<bizarre sup>}
}
\usepackage{siunitx}

\newcommand{\commentout}[1]{}

\definecolor{light-gray}{gray}{0.80}
\newcommand{\tb}[1]{\textbf{#1}}

\newcommand\aref{Alg.~\ref}

\newcommand\eref{Eq.~\ref}
\newcommand\fref{Fig.~\ref}
\newcommand\tref{Tab.~\ref}
\newcommand\sref{\S~\ref}

\newcommand\ha{ \rowcolor{orange!0}}
\newcommand\hb{ \rowcolor{orange!15}}
\newcommand\hc{ \rowcolor{orange!40}}

\usepackage{amsthm}

\usepackage{amsthm}

\newtheorem{mytheorem}{Theorem}
\newtheorem{assumption}[mytheorem]{Assumption}
\newtheorem{lemma}[mytheorem]{Lemma}
\newtheorem{remk}[mytheorem]{Remark}

\usepackage{hyperref}

\usepackage[accepted]{icml2021}

\icmltitlerunning{\OURS: Integer-only BERT Quantization}

\begin{document}
\setlength{\textfloatsep}{6pt}
\newcommand{\OURS}{\textsc{I-BERT}\xspace}

\twocolumn[
\icmltitle{\OURS: Integer-only BERT Quantization}

% It is OKAY to include author information, even for blind
% submissions: the style file will automatically remove it for you
% unless you've provided the [accepted] option to the icml2021
% package.

% List of affiliations: The first argument should be a (short)
% identifier you will use later to specify author affiliations
% Academic affiliations should list Department, University, City, Region, Country
% Industry affiliations should list Company, City, Region, Country

% You can specify symbols, otherwise they are numbered in order.
% Ideally, you should not use this facility. Affiliations will be numbered
% in order of appearance and this is the preferred way.
\icmlsetsymbol{equal}{*}

\begin{icmlauthorlist}
\icmlauthor{Sehoon Kim}{equal,be}
\icmlauthor{Amir Gholami}{equal,be}
\icmlauthor{Zhewei Yao}{equal,be}
\icmlauthor{Michael W. Mahoney}{be}
\icmlauthor{Kurt Keutzer}{be}
\end{icmlauthorlist}

\icmlaffiliation{be}{University of California, Berkeley}

\icmlcorrespondingauthor{Sehoon Kim}{sehoonkim@berkeley.edu}
\icmlcorrespondingauthor{Amir Gholami}{amirgh@berkeley.edu}
\icmlcorrespondingauthor{Zhewei Yao}{zheweiy@berkeley.edu}
\icmlcorrespondingauthor{Michael W. Mahoney}{mahoneymw@berkeley.edu}
\icmlcorrespondingauthor{Kurt Keutzer}{keutzer@berkeley.edu}

% You may provide any keywords that you
% find helpful for describing your paper; these are used to populate
% the "keywords" metadata in the PDF but will not be shown in the document
\icmlkeywords{Machine Learning, ICML}

\vskip 0.3in
]

% this must go after the closing bracket ] following \twocolumn[ ...

% This command actually creates the footnote in the first column
% listing the affiliations and the copyright notice.
% The command takes one argument, which is text to display at the start of the footnote.
% The \icmlEqualContribution command is standard text for equal contribution.
% Remove it (just {}) if you do not need this facility.

% \printAffiliationsAndNotice{}  % leave blank if no need to mention equal contribution
\printAffiliationsAndNotice{\icmlEqualContribution} % otherwise use the standard text.

\begin{abstract}
Transformer based models, like BERT and RoBERTa, have achieved state-of-the-art results in many Natural Language Processing tasks.
However, their memory footprint, inference latency, and power consumption are prohibitive
for efficient inference at the edge, and even at the data center.
While quantization can be a viable solution for this,
previous work on quantizing Transformer based models use floating-point arithmetic during inference, which cannot efficiently utilize integer-only logical units
such as the recent Turing Tensor Cores, or traditional integer-only ARM processors.
In this work, we propose I-BERT, a novel quantization scheme for Transformer based 
models that quantizes the entire inference with integer-only arithmetic.
Based on lightweight integer-only approximation methods for nonlinear operations, e.g., GELU, Softmax, and Layer Normalization, I-BERT performs an end-to-end integer-only BERT inference without any floating point calculation.
We evaluate our approach on GLUE downstream tasks using RoBERTa-Base/Large. 
We show that for both cases, \OURS achieves similar (and slightly higher) accuracy as compared to the full-precision~baseline. 
Furthermore, our preliminary implementation of I-BERT shows a speedup of $2.4- 4.0\times$ for INT8 inference on a T4 GPU system as compared to FP32 inference.
The framework has been developed in PyTorch and has been open-sourced~\cite{kim2021ibert}.

\end{abstract}  
\vspace{-4mm}
\section{\textbf{ Introduction}}
\label{sec:intro}

The recent Transformer based Neural Network (NN) models~\cite{vaswani2017attention}, pre-trained from large unlabeled data (e.g., BERT~\cite{devlin2018bert}, RoBERTa~\cite{liu2019roberta}, and the GPT family~\cite{radford2018improving,radford2019language,brown2020language}), have achieved a significant accuracy improvement
when fine-tuned on a wide range of Natural Language Processing (NLP) tasks 
such as sentence classification~\cite{wang2018glue} and question answering~\cite{rajpurkar2016squad}.   
Despite the state-of-the-art results in various NLP tasks, pre-trained Transformer models are 
generally orders of magnitude larger than prior models.
For example, the BERT-Large model~\cite{devlin2018bert} contains 340M parameters.
Much larger Transformer models have been introduced in the past few years, with even more parameters~\cite{radford2019language, brown2020language, shoeybi2019megatron, rosset2019turing, yang2019xlnet, lepikhin2020gshard, raffel2019exploring}.
Efficient deployment of these models has become a major challenge, even in data centers, due to limited resources (energy, memory footprint, and compute) and the need for real-time inference. 
Obviously, these challenges are greater for edge devices, where the compute and energy
resources are more constrained.

One promising method to tackle this challenge is quantization~\cite{krishnamoorthi2018quantizing, dong2019hawq,wu2018mixed,zhang2018lq,wu2016quantized, jacob2018quantization},
a procedure which compresses NN models into smaller size by representing parameters and/or activations with low bit precision, e.g., 8-bit integer (INT8) instead of 32-bit floating point (FP32). 
Quantization reduces memory footprint by storing parameters/activations in low precision.
With the recent integer-only quantization methods, one can also benefit from faster inference speed by using
low precision integer multiplication and accumulation, instead of floating point arithmetic.
However, previous quantization schemes for Transformer based models use simulated quantization (aka fake quantization),
where all or part of operations in the inference (e.g., GELU~\cite{hendrycks2016gaussian}, Softmax, and Layer Normalization~\cite{ba2016layer}) are carried out with floating point arithmetic~\cite{shen2020q,zafrir2019q8bert,bhandare2019efficient}. 
This approach has multiple drawbacks for deployment in real edge application scenarios.
Most importantly, the resulting NN models cannot be deployed on neural accelerators or popular edge processors that do not support floating point arithmetic.
For instance, the recent server class of Turing Tensor Cores have added
high throughput integer logic that are faster than single/half-precision.
Similarly,
some of the edge processor cores in ARM Cortex-M~\cite{armcortexm} family for embedded systems only contain integer arithmetic units, and they can only support NN deployment with the integer-only kernels~\cite{lai2018cmsis}.
Moreover, one has to consider that compared to the integer-only inference, the approaches that use floating point arithmetic are inferior in latency and power efficiency. 
For chip designers wishing to support BERT-like models, adding floating point arithmetic logic occupies larger die area on a chip, as compared to integer arithmetic logic.
Thus, the complete removal of floating point arithmetic for inference could have a major impact on designing applications, software, and hardware for efficient inference at the edge~\cite{armcortexm}.

While prior work has shown the feasibility of integer-only inference~\cite{jacob2018quantization,yao2020hawqv3}, these approaches have only
focused on models in computer vision with simple CNN layers, Batch Normalization (BatchNorm)~\cite{ioffe2015batch}, and ReLU activations.
These are all linear or piece-wise linear operators.
Due to the non-linear operations used in Transformer architecture, e.g., GELU, Softmax, and Layer Normalization (LayerNorm), these methods cannot be applied to Transformer based models.
Unlike ReLU, computing GELU and Softmax with integer-only arithmetic is not straightforward, due to their non-linearity.
Furthermore, unlike BatchNorm whose parameters/statistics can be fused into the previous convolutional layer in inference,
LayerNorm requires the dynamic computation of the square root of the variance for each  input.
This cannot be na\"ively  computed with integer-only arithmetic.
Another challenge is that processing
%it is known that 
GELU, Softmax, and LayerNorm with low precision can result in signifciant accuracy degradation~\cite{zafrir2019q8bert, bhandare2019efficient}.
For these reasons, other quantization methods such as~\cite{zafrir2019q8bert,shen2020q,bhandare2019efficient} keep these operations in FP32 precision.

In this work, we propose \OURS to address these challenges. 
\OURS incorporates a series of novel integer-only quantization scheme for Transformer based models.
Specifically, our contributions are:
\vspace{-2mm}
\begin{itemize}[noitemsep, nolistsep, labelindent=0pt, leftmargin=*]
    \item 
    We propose new kernels for the efficient and accurate integer-only computation of GELU and Softmax.
    In particular, we approximate GELU and Softmax with light-weight second-order polynomials, which can
    be evaluated with integer-only arithmetic.
    We utilize different techniques to improve the approximation error, and achieve
    a  maximum error of $1.8 \times 10^{-2}$ for GELU, and $1.9 \times 10^{-3}$ for Softmax.
    See~\sref{subsection:gelu} and~\ref{subsection:softmax} for details.
    
    \item 
    For LayerNorm, we perform integer-only computation by leveraging a known algorithm for integer calculation of square root~\cite{crandall2006prime}. 
    See~\sref{subsection:layernorm} for details.
    
    \item 
    We use these approximations of GELU, Softmax, and LayerNorm to design integer-only quantization for Transformer based models. 
    Specifically, we process Embedding and matrix multiplication (MatMul) with INT8 multiplication
    and INT32 accumulation. The following non-linear operations (GELU, Softmax, and LayerNorm)
    are then calculated on the INT32 accumulated result and then requantized back to INT8.
    We represent all parameters and activations in the entire computational graph with integers, and we never cast them into floating point. 
    See ~\fref{fig:overview} (right) for a schematic description. 
    
    \item 
    We apply \OURS to RoBERTa-Base/Large, and we evaluate their accuracy on the GLUE~\cite{wang2018glue} downstream tasks. 
    \OURS achieves similar results as compared to full-precision baseline. Specifically, \OURS outperforms the baseline by 0.3 and 0.5 on the GLUE downstream tasks for RoBERTa-Base and RoBERTa-Large, respectively.
    See \tref{tab:ibert_result} in \sref{subsection:accuracy_eval} for details.
    
    \item
    We deploy INT8 BERT models with the integer-only kernels for non-linear operations on a T4 GPU using TensorRT~\cite{tensorrt}.
    We show that INT8 inference achieves up to 4$\times$ speedup as compared to FP32 inference. 
    See \tref{tab:speedup} in \sref{subsection:latency_eval} for details.
\end{itemize}

\section{Related Work}
\label{sec:relatedwork}
\textbf{Efficient Neural Network.}
There are several different approaches 
to reduce the memory footprint, latency, and power of modern NN 
architectures.
These techniques can be broadly categorized into:
(1) pruning~\cite{han2015learning, li2016pruning, mao2017exploring, lecun1990optimal, molchanov2016pruning, yang2017designing,michel2019sixteen, fan2019reducing, gordon2020compressing, raganato2020fixed, mao2020ladabert, sanh2020movement};
(2) knowledge distillation~\cite{hinton2015distilling, mishra2017apprentice, polino2018model, romero2014fitnets,sanh2019distilbert, sun2019patient, jiao2019tinybert, tang2019distilling, turc2019well, sun2020mobilebert, wang2020minilm, xu2020bert};
(3) efficient neural architecture design~\cite{iandola2016squeezenet, sandler2018mobilenetv2, tan2019efficientnet, howard2019searching, lan2019albert, dehghani2018universal};
(4) hardware-aware NN co-design~\cite{han2017efficient,gholami2018squeezenext, kwon2018co}; and
(5) quantization.

Here, we only focus on quantization and briefly discuss the related work.

%%%%%%%%%%%%%%%%%%%%%%%%%%%%%%%%%%%%%%%%%%%%%%%%%%%%%%%%%%%%%%%%%%%%
\begin{figure*}
\centering{
  \includegraphics[width=0.95\textwidth]{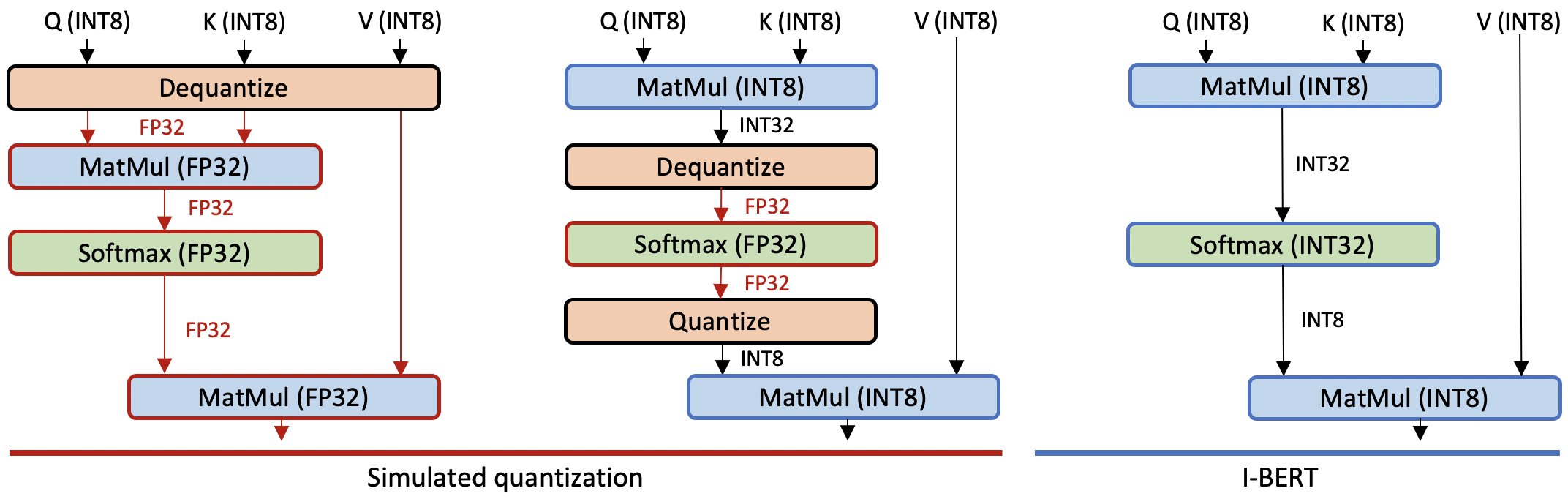}
  \vskip -0.15in
  \caption{Comparison of different quantization schemes applied to the self-attention layer in the Transformer architecture. 
  (Left) Simulated quantization, where all operations are performed with floating point arithmetic. 
  Parameters are quantized and stored as integer, but they are dequantized into floating point for inference. 
  (Middle) Simulated quantization, where only a part of operations are performed with integer arithmetic. 
  Because the Softmax in this figure is performed with floating point arithmetic, the input to the Softmax should be dequantized; and the output from the Softmax should be quantized back into integer to perform the subsequent integer MatMul.
  (Right) The integer-only quantization that we propose. 
  There is neither floating point arithmetic nor dequantization during the entire inference.
  }
  \label{fig:overview}
  }
  \vspace{-3mm}
\end{figure*}
%%%%%%%%%%%%%%%%%%%%%%%%%%%%%%%%%%%%%%%%%%%%%%%%%%%%%%%%%%%%%%%%%%%%

\textbf{Quantization.}
For quantization, the parameters and/or activations are represented with low bit precision~\cite{choi2018pact, courbariaux2015binaryconnect, dong2019hawq, jacob2018quantization, rastegari2016xnor, zhang2018lq, zhou2016dorefa, li2016ternary, wu2016quantized, courbariaux2016binarized, wang2018haq}.
While this line of research mostly focuses on CNN models, there have been recent attempts to introduce quantization techniques into Transformer based models as well.
For example, \cite{bhandare2019efficient} and~\cite{zafrir2019q8bert} propose an 8-bit quantization scheme for Transformer based models and compress the model size up to 25\% of the original size.
Another work~\cite{shen2020q} applies uniform and mixed-precision to quantize BERT model,
where a second-order sensitivity method is used for the mixed-precision setting.
\cite{fan2020training} quantizes a different subset of weights in each training iteration to make models more robust to quantization.
Recently, there have been attempts to quantize BERT with even lower precision. 
\cite{zadeh2020gobo} presents a 3/4-bit centroid-based quantization method that does not require fine-tuning.
\cite{zhang2020ternarybert,bai2020binarybert} leverage knowledge distillation~\cite{hinton2015distilling} to ternarize/binarize weights. 
\cite{jin2021kdlsq} combines knowledge distillation and learned step size quantization~\cite{esser2019learned} method to achieve up to 2-bit quantization of BERT.

However, to the best of our knowledge, all of the prior quantization work on Transformer based models use \textit{simulated quantization} (aka fake quantization), where all or part of operations are performed with floating point arithmetic.
This requires the quantized parameters and/or activations to be dequantized back to FP32 for the floating point operations. 
For example, \cite{shen2020q, zadeh2020gobo} perform the entire inference using floating point arithmetic, as schematically shown in~\fref{fig:overview} (left).
While~\cite{bhandare2019efficient, zafrir2019q8bert, zhang2020ternarybert, bai2020binarybert} attempt to process Embedding and MatMul efficiently with integer arithmetic, they keep the remaining operations (i.e., GELU, Softmax, and LayerNorm) in FP32, as illustrated in~\fref{fig:overview} (middle).
However, our method \OURS uses integer-only quantization for the entire inference process---i.e., without any floating point arithmetic and without any dequantization during the entire inference.
This is illustrated in~\fref{fig:overview} (right). 
This allows more efficient hardware deployment on specialized accelerators or integer-only processors~\cite{armcortexm} as well as faster and less energy consuming inference. 
While we focus on uniform quantization, our method is complementary to other mixed and/or low-precision methods, and can be deployed for those settings as well.

To briefly discuss, there are also several quantization works for computer vision.
\cite{jacob2018quantization} introduces an integer-only quantization scheme for popular CNN models,
by replacing all floating point operations (e.g., convolution, MatMul, and ReLU) with integer operations.
Similarly, the recent work of~\cite{yao2020hawqv3} extends this approach to low precision and mixed precision dyadic quantization, which is an extension of integer-only quantization where no integer division is used.
However, both of these works are limited to CNN models that only contain linear and piece-wise linear operators, and they cannot be applied to Transformer based models with non-linear operators, e.g., GELU, Softmax, and LayerNorm. 
Our work aims to address this limitation by extending the integer-only scheme to the Transformer based models without accuracy~drop.
\section{\textbf{Methodology}}
\label{sec:methodology}
%%%%%%%%%%%%%% BASIC QUANTIZATION METHOD %%%%%%%%%%%%%%%%%%%%%%%%%%

\subsection{\textbf{Basic Quantization Method}}
\label{subsec:basic_notation}

Under \textit{uniform symmetric quantization} scheme, a real number $x$ is
uniformly mapped to an integer value $q \in [-2^{b-1}, 2^{b-1} - 1]$, where
$b$ specifies the quantization bit precision.
The formal definition is:
\begin{equation}
\label{eq:uniform_quantization}
\small
q = \mathrm{Q}(x, b, S) = \mathrm{Int}\bigg(\frac{\mathrm{clip}(x, -\alpha, \alpha)}{S}\bigg),
\end{equation}
where $\mathrm{Q}$ is the quantization operator, $\mathrm{Int}$ is the integer map (e.g., round to the nearest integer), $\mathrm{clip}$ is the truncation function, $\alpha$ is the
clipping parameter used to control the outliers, and $S$ is the scaling factor defined as $\alpha / (2^{b-1} - 1)$.
The reverse mapping from the quantized values $q$ to the real values (aka dequantization) is:
\begin{equation}
\small
\tilde{x} = \mathrm{DQ}(q, S) = Sq \approx x,
\end{equation}
where $\mathrm{DQ}$ denotes the dequantization operator. 
This approach is referred to as uniform symmetric quantization.
It is \textit{uniform} because the spacing between quantized values and their corresponding mapping to real
values is constant. However, several different non-uniform quantization methods have also been proposed~\cite{wu2016quantized, zhang2018lq, choi2018pact, park2018value}.
While non-uniform quantization approaches may better capture the distribution of parameters/activations than uniform quantization, they are in general difficult to deploy on hardware (as they often require a look up table which results in overhead).
Thus, we focus only on uniform quantization in this work.
In addition, this approach is \textit{symmetric} because we clip the values symmetrically within a range $[-\alpha, \alpha]$;
while in asymmetric quantization, the left and right side of this range could be asymmetric/different. 
Finally, we use \textit{static quantization} where all the scaling factors $S$ are fixed during inference to avoid runtime overhead of computing them. 
See~\sref{appendix:quantization_methods} for more details in quantization methods.

%%%%%%%%%%%%%% NON-LINEAR FUNCTIONS %%%%%%%%%%%%%%%%%%%%%%%%%%
\subsection{\textbf{ Non-linear Functions with Integer-only Arithmetic}}

The key to integer-only quantization is to perform all operations with integer arithmetic without using any floating point calculation.
Unlike linear (e.g., MatMul) or piece-wise linear operations (e.g., ReLU), this is not straightforward for non-linear operations (e.g., GELU, Softmax, and LayerNorm). 
This is because the integer-only quantization algorithms in previous works~\cite{yao2020hawqv3, jacob2018quantization} rely on the linear property of the operator.
For example, $\mathrm{MatMul}(Sq)$ is equivalent to $S\cdot \mathrm{MatMul}(q)$ 
for the linear MatMul operation.
This property allows us to apply integer MatMul to the quantized input $q$ and then multiply the scaling factor $S$ to obtain the same result as applying floating point MatMul to the dequantized input $Sq$. 
Importantly, this property \emph{does not} hold for non-linear operations, e.g., $\mathrm{GELU}(Sq) \not=S\cdot \mathrm{GELU}(q)$.
One na\"ive solution is to compute the results of these operations and store them in a look up table~\cite{lai2018cmsis}. 
However, such an approach can have overhead when deployed on chips with limited on-chip memory,
and will create a bottleneck proportional to how fast the look up table could be performed.
Another solution is to dequantize the activations and convert them to floating point, and then compute these non-linear operations with single precision logic~\cite{zafrir2019q8bert, bhandare2019efficient}.
However, this approach is not integer-only and cannot be used on specialized efficient hardware that does not support floating point arithmetic, e.g., ARM Cortex-M~\cite{armcortexm}.

To address this challenge, we approximate non-linear activation functions, GELU and Softmax, with polynomials that can be computed with integer-only arithmetic.
Computing polynomials consists of only addition and multiplication, 
which can be performed with integer arithmetic.
As such, if we can find good polynomial approximations to these operations, then we can perform the entire inference with integer-only
arithmetic. 
For instance, a second-order polynomial represented as $a(x+b)^2+c$ can be efficiently calculated with integer-only arithmetic as shown in~\aref{alg:intpoly}.\footnote{In~\aref{alg:intpoly}, $\floor{\cdot}$ means the floor function. Note that, $q_b$, $q_c$, and $S_{out}$ can be pre-computed under static quantization. That is to say, there is no floating point calculation, e.g., of $S/b$, in inference.}

%%%%%%%%%%%%%%%%%%%%%%%%%%%%%%%%%%%%%%%%%%%%%%%%%%%%%%%%%%%%%%%%%%%%%%%%

\begin{algorithm}[tb]
\caption{\footnotesize
   Integer-only Computation of Second-order Polynomial $a(x + b)^2 + c$}
\label{alg:intpoly}
\small
\begin{algorithmic}
\STATE {\bfseries Input:} $q, S$: quantized input and scaling factor 
\STATE {\bfseries Output:} $q_{out}, S_{out}$: quantized output and scaling factor
\vskip 0.075in
\FUNCTION{\textsc{I-Poly}$(q, S)$ 
\hspace*{\fill}{$\triangleright$ $ qS = x$}
}{  
    \STATE $q_b \leftarrow \floor{b / S}$ 
    \STATE $q_c \leftarrow \floor{c / aS^2}$
    \STATE $S_{out} \leftarrow \floor{aS^2}$
    \STATE $q_{out} \leftarrow (q + q_b)^2 + q_c$ 
    \STATE \Return $q_{out}, S_{out}$
    \hspace*{\fill}{$\triangleright$ 
    $ q_{out}S_{out} \approx  a(x + b)^2 + c\enspace$}

   }
\ENDFUNCTION
\end{algorithmic}
\end{algorithm}

%%%%%%%%%%%%%% POLYNOMIAL APPROXIMATION %%%%%%%%%%%%%%%%%%%%%%%%%%

\subsection{\textbf{Polynomial Approximation of Non-linear Functions}}

There is a large body of work on approximating a function with a polynomial~\cite{stewart1996afternotes}.
We use a class of \textit{interpolating polynomials}, where we are given
the function value for a set of  $n+1$ different data points $\{(x_0, f_0),\dots, (x_n, f_n)\}$, and we seek
to find a polynomial of degree at most $n$ that exactly matches the function value at these points.
It is known that there exists a unique polynomial of degree at most $n$ that passes through all the data points~\cite{waring1779vii}.
We denote this polynomial by $L$, defined as:
\begin{equation}
\label{eqn:lagrange}
\small
L(x) = \sum_{i=0}^{n} f_i l_i(x) \enspace \text{\upshape where} \enspace 
l_i(x) = \prod_{\substack{0 \le j \le n \\ j \ne i}} \frac{x - x_j}{x_i - x_j}.
\end{equation}

Interestingly for our problem, we have two knobs to change to find the best polynomial approximation.
Since we know the actual target function and can query its exact value
for any input,
we can choose the interpolating point $(x_i, f_i)$ to be any point on the function.
The second knob is to choose the degree of the polynomial.
While choosing a high-order polynomial results in smaller error (see Appendix~\ref{sec:error_of_lagrange}), there are two problems with this.
First, high-order polynomials have higher computational and memory overhead.
Second, it is challenging to evaluate
them with low-precision integer-only arithmetic,
as overflow can happen when multiplying integer values. For every multiplication, we need
to use double bit-precision to avoid overflow.
As such, the challenge is to find a good low-order polynomial that can closely approximate the
non-linear functions used in Transformers. 
This is what we discuss next, for GELU and Softmax, in~\sref{subsection:gelu} and~\ref{subsection:softmax}, respectively, where we show that one can
get a close approximation by using only a second-order~polynomial.

%%%%%%%%%%%%%%%%%%%%%%%%%%%% GELU %%%%%%%%%%%%%%%%%%%%%%%%%%%%%%%%%%
\subsection{\textbf{Integer-only GELU}}
\label{subsection:gelu}

\begin{figure}[]
\centering{
\centerline{
  \includegraphics[width=0.48\textwidth]{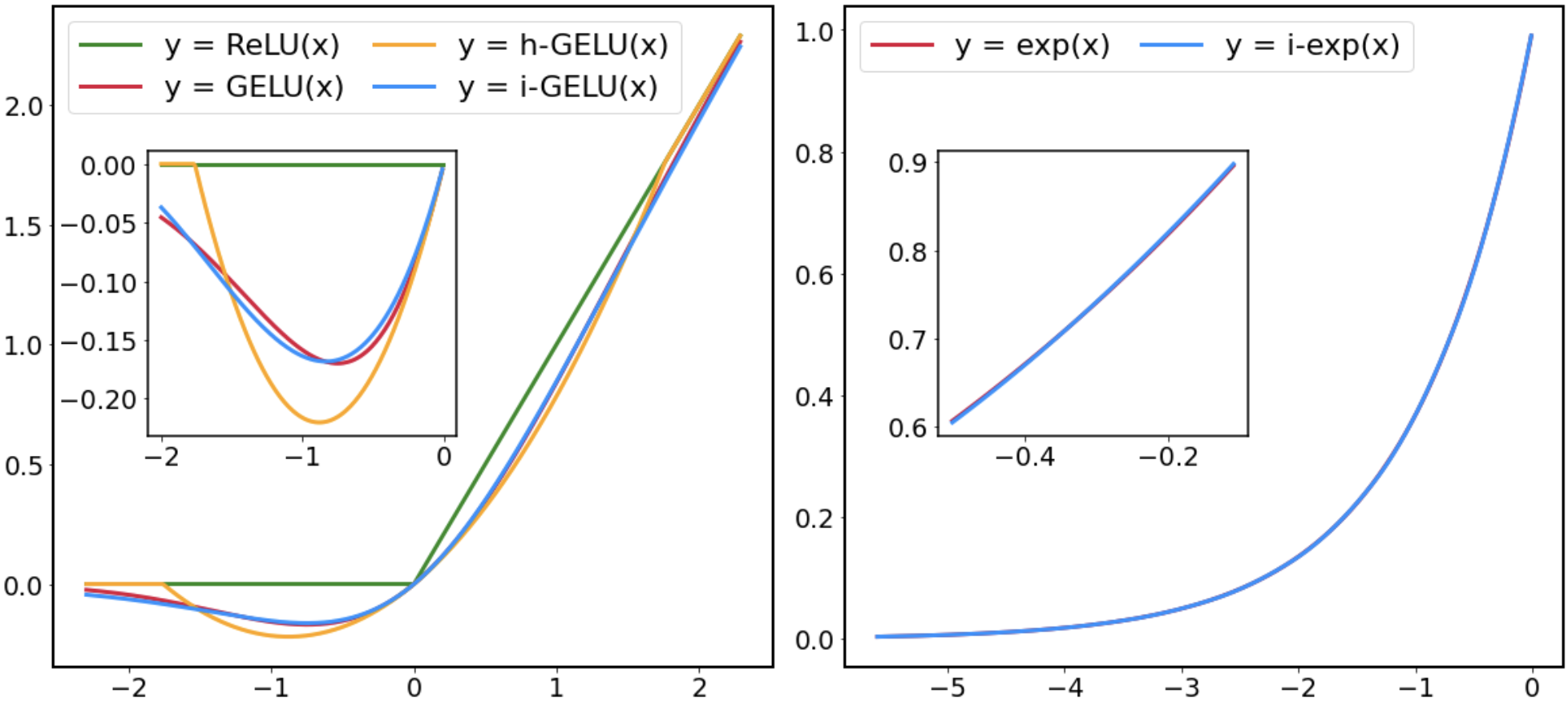}
  }
  \caption{
  (Left) Comparison between RELU, GELU, h-GELU and i-GELU. 
  (Right) Comparison between exponential (exp) and our integer-only exponential (i-exp). 
  }
  \label{fig:gelu-exp}
  }
  \vspace{1mm}
\end{figure}

GELU~\cite{hendrycks2016gaussian} is a non-linear activation function used in Transformer models, defined~as:
\begin{equation}
\small
\begin{split}
\label{eq:gelu}
    \mathrm{GELU}(x) &:= x \cdot \frac12 \left[ 1 + \mathrm{erf}(\frac{x}{\sqrt{2}})\right], \\
    \text{where} \enspace \mathrm{erf}(x) &:= \frac{2}{\sqrt{\pi}}\int_0^x \exp{(-t^2)} dt.
\end{split}
\end{equation}
Here, $\mathrm{erf}$ is the error function. Figure~\ref{fig:gelu-exp} shows the behaviour of the GELU function (shown in red).
GELU has a similar behaviour as ReLU (shown in green) in the limit of large positive/negative values,
but it behaves differently near zero.
Direct evaluation of the integration term in $\mathrm{erf}$ is not computationally efficient.
For this reason, several different approximations have been
proposed for evaluating GELU. For example, \cite{hendrycks2016gaussian} suggests using
Sigmoid to approximate $\mathrm{erf}$:
\begin{equation}
\label{eqn:gelu-approx}
\small
\mathrm{GELU}(x) \approx x \sigma(1.702 x),
\end{equation}
where $\sigma(\cdot)$ is the Sigmoid function.
This approximation, however, is not a viable solution for integer-only quantization, as the Sigmoid itself is another non-linear function which requires
floating point arithmetic.
One way to address this is to approximate Sigmoid with the so-called
hard Sigmoid (h-Sigmoid) proposed by~\cite{howard2019searching} (designed in the context of efficient computer vision models) to obtain an integer-only approximation for GELU:
\begin{equation}
\small
\label{eqn:hgelu}
\mathrm{h{\text-}GELU}(x) := x\frac{\mathrm{ReLU6}(1.702x+3)}{6} \approx \mathrm{GELU}(x).
\end{equation} 
We refer to this approximation as h-GELU.
Although h-GELU can be computed with integer arithmetic, we observed that replacing GELU with h-GELU in Transformers results in a significant accuracy drop.
This is due to the large gap between h-GELU and GELU as depicted in~\tref{tab:sigmoid-approximation}.%
\footnote{Later in our ablation study, we show this can lead to accuracy degradation of up to 2.2 percentages, as reported in~\tref{tab:gelu_comparison}.}
Figure~\ref{fig:gelu-exp} (left) also shows the noticeable gap between those two functions.

A simple way to address the above problem is to use polynomials to approximate GELU, by solving
the following optimization problem:
\begin{equation}
\small
\begin{split}
\label{eq:gelu_opt_objective}
    &\min_{a,b,c} \frac12\left\| \mathrm{GELU}(x) 
    - x \cdot \frac{1}{2} \left[ 1 + \mathrm{L}(\frac{x}{\sqrt{2}})\right]\right\|_2^2 , \\
    &\mathrm{s.t.} \quad L(x) = a(x+b)^2 + c,
\end{split}
\end{equation}
where $\mathrm{L}(x)$ is a second-order polynomial used to approximate the $\mathrm{erf}$ function. 
Directly optimizing~\eref{eq:gelu_opt_objective} results in a poor approximation since the definition domain of $\mathrm{erf}$ contains the entire real numbers. 
To address this, we only optimize $L(x)$ in a limited range since $\mathrm{erf}$ approaches to 1 ($-$1) for large values of $x$.
We also take advantage of the fact that $\mathrm{erf}$ is an odd function (i.e., $\mathrm{erf}(-x)=-\mathrm{erf}(x)$), and thus only consider approximating it in the positive
domain.
After finding the best interpolating points, i.e., $(x_i,f_i)$ in~\eref{eqn:lagrange}, and applying these adjustments we arrive at the following polynomial:
\begin{equation}
\label{eq:bestL_erf}
\small
    \mathrm{L}(x) = \mathrm{sgn}(x)\left[a(\mathrm{clip}(|x|, \max=-b)+b)^2 + 1\right],
\end{equation}
where $a=-0.2888$ and $b = -1.769$, and $\mathrm{sgn}$ denotes the sign function.
\footnote{Note that $L(x)$ is approximating GELU in the range of $[0, -b]$.}
Using this polynomial we arrive at i-GELU, the integer-only approximation for GELU, defined~as:
\begin{equation}
\label{eqn:igelu}
\small
\mathrm{i{\text-}GELU}(x) := x \cdot \frac{1}{2} \left[ 1 + \mathrm{L}(\frac{x}{\sqrt{2}})\right].
\end{equation}

%%%%%%%%%%%%%%%%%%%%%%%%%%%%%%%%%%%%%%%%%%%%%%%%%%%%%%%%%%%%%%%
%%%%%%%%%%%%%%%%%%%%%%%%%%%%%%%%%%%%%%%%%%%%%%%%%%%%%%%%%%%%%%%

Algorithm~\ref{alg:intgelu} summarizes the integer-only computation of GELU using i-GELU.
We illustrate the behaviour of i-GELU in~\fref{fig:gelu-exp} (left). As one
can see, i-GELU closely approximates GELU, particularly around the origin.
We also report the approximation error of i-GELU along with h-GELU
in~\tref{tab:sigmoid-approximation}, where i-GELU has an
average error of $8.2 \times 10^{-3}$ and a maximum error of $1.8 \times 10^{-2}$.
This is $\sim3\times$ more accurate than h-GELU whose average and maximum errors are $3.1 \times 10^{-2}$ and $6.8 \times 10^{-2}$, respectively.
Also, i-GELU even slightly outperforms the Sigmoid based approximation of~\eref{eqn:gelu-approx}, but without using any floating point
arithmetic. 
Note that computing the Sigmoid requires floating~point.
Later in the results section, we show that this improved approximation, actually
results in better accuracy of i-GELU as compared to h-GELU (see~\tref{tab:gelu_comparison}).

%%%%%%%%%%%%%%%%%%%%%%%%%%%%%%%%%%%%%%%%%%%%%%%%%%%%%%%%%%%%%%%%%%%%%%%%

\begin{algorithm}[tb]
\caption{\footnotesize
Integer-only GELU }
\label{alg:intgelu}
\small
\begin{algorithmic}
\STATE {\bfseries Input:} $q, S$: quantized input and scaling factor  
\STATE {\bfseries Output:} $q_{out}, S_{out}$: quantized output and scaling factor 
\vskip 0.075in
\FUNCTION{\textsc{I-Erf}$(q, S)$
\hspace*{\fill}{$\triangleright$ $ qS = x$}
}{
    \STATE $a, b, c \leftarrow -0.2888, -1.769, 1$
    \STATE $q_{\mathrm{sgn}}, q 
    \leftarrow \mathrm{sgn}(q), \mathrm{clip}(|q|, \mathrm{max} = -b/S)$
    \STATE $q_{L}, S_{L} \leftarrow$ \textsc{I-Poly}${(q, S)}$ with $a, b, c$
    \hspace*{\fill}{$\triangleright$ \eref{eq:bestL_erf}$\enspace$}  
    \STATE $q_{out}, S_{out} \leftarrow q_{\mathrm{sgn}}q_L, S_L$ 
    \STATE \Return $q_{out}, S_{out}$
    \hspace*{\fill}{$\triangleright$ 
    $ q_{out}S_{out} \approx \mathrm{erf}(x)\enspace$}
   }
\ENDFUNCTION
 
\vskip 0.075in
\FUNCTION{\textsc{I-Gelu}$(q, S)$
\hspace*{\fill}{$\triangleright$ $ qS = x$}
}{
    \STATE $q_{\mathrm{erf}}, S_{\mathrm{erf}} \leftarrow$ \textsc{I-Erf}${(q, S / \sqrt{2})}$ 
    \STATE $q_{1} \leftarrow \floor{1/S_{\mathrm{erf}}}$
    \STATE $q_{out}, S_{out} 
    \leftarrow q(q_{\mathrm{erf}} + q_{1}), SS_{\mathrm{erf}}/2$
    \label{algline:geluapprox}
    \STATE \Return $q_{out}, S_{out}$
    \hspace*{\fill}{$\triangleright$ 
    $ q_{out}S_{out} \approx \mathrm{GELU}(x)\enspace$}
   }
\ENDFUNCTION
\end{algorithmic}
\end{algorithm}

%%%%%%%%%%%%%%%%%%%%%%%%%%%%%%%%%%%%%%%%%%%%%%%%%%%%%%%%%%%%%%%
\begin{table}[!t]
\vspace{-2mm}
\caption{ 
Comparison of different approximation methods for GELU. 
The second column (Int-only) indicates whether each approximation method can be computed with integer-only arithmetic.
As metrics for approximation error, we report L$^2$ and L$^\infty$ distance from GELU across the range of [-4, 4]. 
}

\vskip 0.1in
\label{tab:sigmoid-approximation}
\centering
\centerline{
\small{
\begin{tabular}{lccc}
\toprule
\ha                     & Int-only   & L$^2$ dist  & L$^\infty$ dist\\ 
\midrule              
\ha $x \sigma(1.702x)$  & \xmark              & 0.012                & 0.020                 \\
\ha h-GELU                & \cmark              & 0.031                & 0.068                 \\
\midrule                                              
\hc i-GELU (Ours)       & \cmark              & \bf{0.0082}          & \bf{0.018}            \\
\bottomrule
\end{tabular} 
}
}
\vspace{2mm}
\end{table}
%%%%%%%%%%%%%%%%%%%%%%%%%%%%%%%%%%%%%%%%%%%%%%%%%%%%%%%%%%%%%%%

%%%%%%%%%%%%%%%%%%%%%%%%%%%%%%%% SOFTMAX %%%%%%%%%%%%%%%%%%%%%%%%%%%%%%%%%%%%%%
\subsection{\textbf{Integer-only Softmax}}
\label{subsection:softmax}

Softmax normalizes an input vector and maps it to 
a probability distribution: 
\begin{equation}
\label{eqn:softmax} 
\small
\mathrm{Softmax}(\mathbf{x})_i := \frac{\exp{x_i}}{\sum_{j=1}^k \exp{x_j}} ,
\enspace \text{where} \enspace \mathbf{x} = [x_1, \dots, x_k].
\end{equation}
Approximating the
Softmax layer with integer arithmetic is quite challenging, as the exponential
function used in Softmax is unbounded and changes rapidly.
As such, prior Transformer quantization techniques~\cite{bhandare2019efficient,zafrir2019q8bert} treat this layer using floating point arithmetic.
Some prior work have proposed look up tables with interpolation~\cite{schraudolph1999fast},
but as before we avoid look up tables and strive for a pure arithmetic based approximation.
In addition, although \cite{hauser2001approximating} proposes polynomial approximation 
methods for the exponential function,
it uses significantly high-degree polynomials, and is only applicable on a limited finite domain. 

Similar to GELU, we cannot use a high-order polynomial, but even using such polynomial
is ineffective to approximate the exponential function in Softmax.
However, it is possible to address problem by limiting the approximation range of Softmax.
First, we subtract the maximum value from the input to the exponential for numerical stability:
\begin{equation}
\label{eqn:softmax2} 
\small
\mathrm{Softmax}(\mathbf{x})_i = \frac{\exp{(x_i - x_{\max})}}{\sum_{j=1}^k \exp{(x_j - x_{\max})}},
\end{equation}
where $x_{\max} = \max_i(x_i)$.
Note that now all the inputs to the exponential function, i.e., $\tilde x_i = x_i - x_{\max}$, become non-positive.
We can decompose any 
non-positive real number $\tilde x$ as $\tilde x = (-\ln{2})z + p$,
where the quotient $z$ is a non-negative integer and the remainder $p$ is a real number in $(-\ln2, 0]$. 
Then, the exponential of $\tilde x$ can be written as:
\begin{equation}
\label{eqn:expshift}
\small
\exp(\tilde x) = 2^{-z} \exp(p) = \exp(p) \verb|>>| z,
\end{equation}
where \verb|>>| is the bit shifting operation.
As a result, we only need to approximate the exponential function in the compact interval of $p\in(-\ln2, 0]$.
This is a much smaller range as compared to the domain of all real numbers.
Interestingly, a variant of this method was used in the Itanium 2 machine from HP~\cite{thomas2004libm,detrey2005parameterized}, but with a look up table
for evaluating $\exp(p)$.

We use a second-order polynomial to approximate the exponential
function in this range. To find the coefficients of the polynomial, we minimize the  L$^2$ distance from exponential function in the interval of $(-\ln2, 0]$. 
This results in the following approximation:
\begin{equation}
\label{eq:bestL_exp}
\small
L(p) = 0.3585 (p + 1.353)^2 + 0.344 \approx \exp(p). 
\end{equation}
Substituting the exponential term in~\eref{eqn:expshift} with this polynomial results in i-exp:
\begin{equation}
\label{eqn:iexp}
\mathrm{i\text{-}exp}(\tilde x) := L(p) \verb|>>| z  
\end{equation}
where $\enspace z = \floor{-\tilde x / \ln 2}$ and $p = \tilde x + z \ln 2$. This can be calculated with integer arithmetic.
Algorithm~\ref{alg:intexp} describes the integer-only computation of the Softmax fucntion using i-exp.
Figure~\ref{fig:gelu-exp} (right) plots the result of i-exp, 
which is nearly identical to the exponential function.
We find that the largest gap between these two functions is only $1.9 \times 10^{-3}$.
Considering that 8-bit quantization of a unit interval introduces a quantization error of $1/256 = 3.9 \times 10^{-3}$, our approximation error is relatively negligible and can be subsumed into the quantization error. 

%%%%%%%%%%%%%%%%%%%%%%%%%%%%%%%%%%%%%%%%%%%%%%%%%%%%%%%%%%%%%%%
\begin{algorithm}[tb]
\caption{\footnotesize
Integer-only Exponential and Softmax}
\label{alg:intexp}
\small
\begin{algorithmic}
\STATE {\bfseries Input:} $q, S$: quantized input and scaling factor  
\STATE {\bfseries Output:} $q_{out}, S_{out}$: quantized output and scaling factor
 
\vskip 0.075in
\FUNCTION{\textsc{I-Exp}$(q, S)$
\hspace*{\fill}{$\triangleright$ $ qS = x$}
}{
    \STATE $a, b, c \leftarrow 0.3585, 1.353, 0.344$
    \STATE $q_{\ln 2} \leftarrow \floor{\ln 2 / S}$
    \STATE $z \leftarrow \floor{-q / q_{\ln 2} }$
    \STATE $q_p \leftarrow q + z q_{\ln 2}$
    \hspace*{\fill}{$\triangleright$ $ q_pS = p\enspace$}
    \STATE $q_L, S_L \leftarrow$ \textsc{I-Poly}${(q_p, S)}$ with $a, b, c$
    \hspace*{\fill}{$\triangleright$ \eref{eq:bestL_exp}$\enspace$}
    \STATE $q_{out}, S_{out} \leftarrow q_L \verb|>>| z, S_L $
    \STATE \Return $q_{out}, S_{out}$    
    \hspace*{\fill}{$\triangleright$ 
    $ q_{out}S_{out} \approx \mathrm{exp}(x)\enspace$}
   }
\ENDFUNCTION
 
\vskip 0.075in
\FUNCTION{\textsc{I-Softmax}$(q, S)$
\hspace*{\fill}{$\triangleright$ $ qS = x$}
}{
    \STATE $\tilde q \leftarrow q - \mathrm{max}(q)$
    \STATE $q_{\mathrm{exp}}, S_{\mathrm{exp}} \leftarrow$ \textsc{I-Exp}${(\tilde q, S)}$  
    \STATE $q_{out}, S_{out} \leftarrow q_{\mathrm{exp}} / \mathrm{sum}(q_{\mathrm{exp}}), S_{\mathrm{exp}}$
    \STATE \Return $q_{out}, S_{out}$    
    \hspace*{\fill}{$\triangleright$ 
    $ q_{out}S_{out} \approx \mathrm{Softmax}(x)\enspace$}
   }
\ENDFUNCTION
\end{algorithmic}
\end{algorithm}

%%%%%%%%%%%%%%%%%%%%%%%%%%%%%%% LayerNorm %%%%%%%%%%%%%%%%%%%%%%%%%%%%%%%%%%%%%
\subsection{\textbf{Integer-only LayerNorm}}
\label{subsection:layernorm}

LayerNorm is commonly used in Transformers and involves several non-linear operations, such as division, square, and square root. 
This operation is used for normalizing the input activation across the channel dimension.
The normalization process is described as:
\begin{equation}
\label{eqn:lnmusigma}
\small
    \tilde{x} = \frac{x - \mu}{\sigma}  \enspace \text{\upshape where} \enspace
    \mu   =  \frac{1}{C}\sum_{i=1}^C x_i 
    \enspace \text{\upshape and} \enspace 
    \sigma = \sqrt{\frac{1}{C}\sum_{i=1}^C (x_i - \mu)^2}  .
\end{equation} 
Here, $\mu$ and $\sigma$ are the mean and standard deviation of the input across the channel dimension.
One subtle challenge here is that the input statistics (i.e., $\mu$ and $\sigma$) change
rapidly for NLP tasks, and these values need to be 
calculated dynamically during runtime. While computing
$\mu$ is straightforward, evaluating $\sigma$ requires the square-root
function.

The square-root function can be efficiently evaluated with integer-only
arithmetic through an iterative algorithm proposed in~\cite{crandall2006prime}, as described in~\aref{alg:intsqrt}.
Given any non-negative integer input $n$, this algorithm iteratively searches for the exact value of $\floor{\sqrt{n}}$ based on Newton's Method and only requires integer arithmetic.
This algorithm is computationally lightweight, as it converges within at most
four iterations for any INT32 inputs and each iteration consists only of one integer division, one integer addition, and one bit-shifting operation. 
The rest of the the non-linear operations in LayerNorm such as division and square are straightforwardly computed with integer arithmetic.

%%%%%%%%%%%%%%%%%%%%%%%%%%%%%%%%%%%%%%%%%%%%%%%%%%%%%%%%%%%%%%%
\begin{algorithm}[tb]
\caption{\footnotesize
Integer-only Square Root }
\label{alg:intsqrt}
\small
\begin{algorithmic}
\STATE {\bfseries Input:} $n$: input integer 
\STATE {\bfseries Output:} integer square root of $n$, i.e., $\floor{\sqrt{n}}$
 
\vskip 0.075in
\FUNCTION{\textsc{I-Sqrt}$(n)$}{
    \STATE \textbf{if} $n=0$ \textbf{then return} $0$
    \STATE Intialize $x_0$ to $2^{\ceil{Bits(n)/2}}$ and $i$ to $0$
    \STATE \textbf{repeat}
    \STATE \hspace{10pt} $x_{i+1} \leftarrow \floor{(x_i + \floor{n / x_i}) / 2}$
    \STATE \hspace{10pt} \textbf{if} $x_{i+1} \ge x_i$  \textbf{then return} $x_i$
    \STATE \hspace{10pt} \textbf{else} $i \leftarrow i+1$
   }
\ENDFUNCTION
\end{algorithmic}
\end{algorithm} 
\section{\textbf{Results}}
\label{sec:results}

%%%%%%%%%%%%%%%%%%%%%%%%%%%%%%%%%%%%%%%%%%%%%%%%%%%%%%%%%%%%%%%%%%%%%%%%%%%%%%
\begin{table*}[!t]
\caption{ 
Integer-only quantization result for RoBERTa-Base and RoBERTa-Large on the development set of the GLUE benchmark. 
Baseline is trained by the authors from the pre-trained models, and \OURS is quantized and fine-tuned from the baseline.
We also report the difference (Diff) between the baseline accuracy and the \OURS accuracy. 
}
\vskip 0.05in
\label{tab:ibert_result}
    \centerline{
\subtable[RoBERTa-Base]
    {
    \centering
    \centerline{
    \small{
    \setlength{\tabcolsep}{5pt}{
       \begin{tabular}{lcccccccccccc}
        \toprule
        \ha             & {Precision}  & {Int-only} & {MNLI-m} & {MNLI-mm} & {QQP }  & {QNLI} & {SST-2} &  {CoLA}    & {STS-B}  & {MRPC}    & {RTE}   & {Avg.}        \\ 
        \midrule         
        \ha Baseline    & FP32              & \xmark            & \tb{87.8}  & \tb{87.4}    & \tb{90.4}  &\tb{92.8}  & 94.6       & 61.2          & \tb{91.1}  & 90.9         &  78.0      &  86.0     \\
        \hc \OURS       & INT8              & \cmark            & 87.5       & \tb {87.4}   & 90.2       & \tb{92.8} & \tb{95.2}  &\tb{62.5}      & 90.8       &\tb{91.1}     & \tb{79.4}  &  \tb{86.3} \\
        \midrule         
        \ha Diff        &                   &                   &  -0.3      &  0.0         & -0.2       & 0.0       & +0.6       & +1.3          & -0.3       &  +0.2        & +1.4       &  +0.3  \\
        \bottomrule
        \end{tabular} 
       
        }
        }
        }
    }
}
    
\centerline{
\subtable[RoBERTa-Large]
    {
    \centering
    \centerline{
    \small{
    \setlength{\tabcolsep}{5pt}{
    
        \begin{tabular}{lccccccccccccc}
        \toprule
        \ha             & {Precision} & {Int-only} & {MNLI-m} & {MNLI-mm} & {QQP }  & {QNLI}  & {SST-2}   &    {CoLA}  & {STS-B}   & {MRPC}    & {RTE}   & {Avg.} \\ 
        \midrule             
        \ha Baseline    & FP32              & \xmark            & 90.0        & 89.9         & 92.8       & 94.1      & 96.3          & 68.0          & \tb{92.2}   & 91.8         & 86.3       & 89.0 \\
        \hc \OURS       & INT8              & \cmark            & \tb{90.4}   & \tb{90.3}    & \tb{93.0}  & \tb{94.5} & \tb{96.4}     &\tb{69.0}      &  \tb{92.2}  & \tb{93.0}    & \tb{87.0}  &  \tb{89.5} \\
        \midrule             
        \ha Diff        &                   &                   & +0.4        & +0.4         & +0.2         & +0.4      & +0.1        &+1.0           & 0.0         & +1.2         & +0.7         & +0.5 \\
        \bottomrule
        \end{tabular}

        % \label{fig:third_sub}
        }
        }
        }
    }
}

\vspace{-6mm}
\end{table*}
%%%%%%%%%%%%%%%%%%%%%%%%%%%%%%%%%%%%%%%%%%%%%%%%%%%%%%%%%%%%%%%%%%%%%%%%%%%%%%

%%%%%%%%%%%%%%%%%%%%%%%%%%%%%%%%%%%%%%%%%%%%%%%%%%%%%%%%%%%%%%%%%%%%%%%%%%%%%%%%%%%

In this section, we first measure the accuracy of \OURS using the General Language Understanding Evaluation~\cite{wang2018glue} (GLUE) benchmark (\sref{subsection:accuracy_eval}).
Then, we discuss the latency speedup of \OURS using direct hardware deployment and compare it with pure FP32 model (\sref{subsection:latency_eval}). 
Finally, we conduct ablation studies to showcase the effectiveness of our integer-only approximation methods (\sref{subsection:ablation_studies}).

% -------------------------------------------------------
\subsection{\textbf{Accuracy Evaluation on GLUE}}
\label{subsection:accuracy_eval}

We implement \OURS on the RoBERTa~\cite{liu2019roberta} model using~\cite{ott2019fairseq}.
For the integer-only implementation, we replace all the floating point operations in the original
model with the corresponding integer-only operations that were discussed in~\sref{sec:methodology}.
In particular, we perform MatMul and Embedding with INT8 precision, and the non-linear operations with INT32 precision, as using INT32 for computing these operations has little overhead. 
See \sref{appendix:implementation_details} for implementation details.
For each of the GLUE downstream tasks, we train both FP32 baseline and integer-only I-BERT models, and evaluate the accuracy on the development set.
See Appendix~\ref{appendix:training_details} and \ref{appendix:accuracy_eval} for training and evaluation details.
While we only test RoBERTa-Base/Large, our method is not restricted to RoBERTa.
The integer-only approximations can be performed for any NN models including Transformers that uses similar non-linear operations.

% -----------------------------------------

% -----------------------------------------
The integer-only quantization results for RoBERTa-Base/Large are presented in~\tref{tab:ibert_result}. 
As one can see, \OURS consistently achieves comparable or slightly higher accuracy than baseline. 
For RoBERTa-Base, \OURS
achieves higher accuracy for all cases (up to 1.4 for RTE), except for MNLI-m, QQP, and STS-B tasks, where we observe a small
accuracy degradation up to 0.3.
We observe a similar behaviour on the RoBERTa-Large model, where \OURS matches or outperforms the baseline accuracy for all the downstream tasks.
On average, \OURS outperforms the baseline by 0.3/0.5 for RoBERTa-Base/Large, respectively.

%%%%%%%%%%%%%%%%%%%%%%%%%%%%%%%%%%%%%%%%%%%%%%%%%%%%%%%%%%%%%%%%%%%%%%%%%%%%%%%%%%%%%%%%%%%%%%%%%%%%%%%%%%%%

\subsection{\textbf{Latency Evaluation}}
\label{subsection:latency_eval}

We evaluate the latency speedup of INT8 inference of \OURS, by direct deployment on
a Tesla T4 GPU with Turing Tensor Cores that supports accelerated INT8 execution.
Although T4 GPU is not a pure integer-only hardware, we select it as our target device due to its extensive software support~\cite{tensorrt, chen2018tvm}, and in
particular Nvidia's TensorRT library~\cite{tensorrt}.
Furthermore, as we do not exploit any T4-specific exclusive features or requirements, our work can be extensively deployed on other hardware as well.
See \sref{appendix:env_setup} for the detailed environment setup.
For evaluation, we implement two variants of BERT-Base/Large: (1) pure FP32 models using na\"ive FP32 kernels for non-linear operations; and (2) quantized INT8 models using customized kernels for the non-linear operations.
The customized kernels compute GELU, Softmax, and LayerNorm based on the integer-only methods described in~\sref{sec:methodology}.
We measure the inference latency for different sequence lengths (128 and 256) and batch sizes (1, 2, 4, and 8). 
    
Table~\ref{tab:speedup} shows the inference latency speedup of INT8 models with respect to FP32 models.
As one can see, the INT8 inference of \OURS is on average 3.08$\times$ and 3.56$\times$ faster than pure FP32 inference for BERT-Base and BERT-Large, respectively, achieving up to 4.00$\times$ speedup.
The result implies that, when deployed on specialized hardware that supports efficient integer computations, \OURS can achieve significant speedup as compared to FP32 models.
Further speedups are possible with NVIDIA's custom Transformer plugins~\cite{tensorrtbert} which fuse
the multi-head attention and Softmax layers (see ~\sref{appendix:env_setup}).

While the greatest value of our work will become evident when 
our approach enables quantization on lower-end microprocessors
without floating-point hardware, 
this demonstration must wait for improved software
support for implementing quantized NN models on those processors.
In the meantime, we believe the promise of our approach is
illustrated by these latency reductions shown above.

%%%%%%%%%%%%%%%%%%%%%%%%%%%%%%%%%%%%%%%%%%%%%%%%%%%%%%%%%%%%%%%%%%%%%%%
\begin{table}[!t]
\caption{ 
Inference latency speedup of INT8 inference with respect to FP32 inference for BERT-Base and BERT-Large. 
Latency is measured for different sentence lengths (SL) and batch sizes (BS). 
}

\vskip 0.1in
\label{tab:speedup}

    \centering
    \small{
    \setlength{\tabcolsep}{3.5pt}{
    \centerline{
      \begin{tabular}{l|cccc|cccc|c}
        \toprule
        \ha   SL           & \multicolumn{4}{c|}{128}    & \multicolumn{4}{c|}{256}      &\multirow{2}{*}[-1.5pt]{Avg.} \\ 
        \ha   BS  &  1      &   2     &  4  & 8  &   1      &   2     &  4  & 8      \\
        \midrule
        \hb Base &  {2.42}  &  {3.36}  &  {3.39}  & {3.31}   &   {3.11}  &   {2.96}  &  {2.94} &  {3.15} & {3.08}   \\
        \hc Large &  {3.20}  &  {4.00}  &  {3.98}  &   {3.81} &   {3.19}  &   {3.51}  &  {3.37} & {3.40}   & {3.56}   \\
        \bottomrule
        \end{tabular} 
    }
    }
    }

\vspace{3mm}
\end{table}
%%%%%%%%%%%%%%%%%%%%%%%%%%%%%%%%%%%%%%%%%%%%%%%%%%%%%%%%%%%%%%%%%%%%%%%

\subsection{\textbf{Ablation Studies}}
\label{subsection:ablation_studies}

% --------------------------------------------------------
Here, we perform an ablation study to show the benefit of i-GELU as compared to
other approximation methods for GELU, and in particular h-GELU in ~\eref{eqn:hgelu}. 
For comparison, we implement two variants of \OURS by replacing i-GELU with GELU and h-GELU, respectively.
The former is the exact computation of GELU with floating point arithmetic, 
and the later is another integer-only approximation method for GELU (see~\sref{sec:methodology}).
We use RoBERTa-Large model as baseline along with the QNLI, SST-2, MPRC, and RTE tasks.
All models are trained and fine-tuned according to the procedure described in~\sref{subsection:accuracy_eval}, and the final accuracies are reported in~\tref{tab:gelu_comparison}.

As one can see, replacing GELU with h-GELU approximation results in accuracy degradation for all downstream tasks except for MRPC.
Accuracy drops by 0.5 on average and up to 1.1 for RTE task.
Although accuracy slightly improves for MRPC, the amount of increase is smaller than replacing GELU with i-GELU.
This empirically demonstrates that h-GELU is not sufficiently tight enough to approximate GELU well.
Approximating GELU with i-GELU results in strictly better accuracy for all four downstream tasks than h-GELU.
In particular, i-GELU outperforms h-GELU by 0.7 on average, and it achieves comparable or slightly better result to the non-approximated full-precision GELU. 
i-GELU also performs better than GELU, which is quite interesting, but at this
time, we do not have an explanation for this behaviour.

%%%%%%%%%%%%%%%%%%%%%%%%%%%%%%%%%%%%%%%%%%%%%%%%%%%%%%%%%%%%%%%%%%%%%%%
\begin{table}[t]
\caption{ 
Accuracy of models that use GELU, h-GELU and i-GELU for GELU computation. Note that the former is full-precision, floating point computation while the latter two are integer-only approximations. 
}
\vskip 0.1in
\label{tab:gelu_comparison}
    \centering
    \small{
    \setlength{\tabcolsep}{5pt}{
    \centerline{
       \begin{tabular}{lccccccc}
        \toprule
        \ha         & Int-only & QNLI & SST-2    & MRPC     & RTE  & Avg.    \\ 
        \midrule
        \ha GELU    & \xmark            & 94.4      & 96.3          & 92.6          & 85.9      & 92.3 \\
        \ha h-GELU  & \cmark            & 94.3      & 96.0          & 92.8          & 84.8      & 92.0 \\
        \midrule
        \hc i-GELU  & \cmark            & \tb{94.5} & \tb{96.4}     & \tb{93.0}     & \tb{87.0} &  \tb{92.7} \\
        \bottomrule
        \end{tabular} 
    }
    }
    }
\vspace{3mm}
\end{table}
%%%%%%%%%%%%%%%%%%%%%%%%%%%%%%%%%%%%%%%%%%%%%%%%%%%%%%%%%%%%%%%%%%%%%%%
%\input{_s5_discussion.tex}
\section{\textbf{Conclusions}}
\label{sec:conclusions}

We have proposed \OURS, a novel integer-only quantization scheme for Transformers, where the entire inference is performed with pure integer arithmetic.
Key elements of \OURS are approximation methods for nonlinear operations such as GELU, Softmax, and LayerNorm, which enable their approximation with integer computation.
We empirically evaluated \OURS on RoBERTa-Base/Large models, where our quantization method improves
the average GLUE score by 0.3/0.5 points as comapred to baseline.
Furthermore, we directly deployed the quantized models and measured the end-to-end inference latency, showing that \OURS can achieve up to 4.00$\times$ speedup on a Tesla T4 GPU as compared to floating point baseline.
As part of future work, one could consider
using our approximation to improve the training speed as well. For instance, one
could consider replacing GELU with i-GELU during training. Also,
further studies are needed to evaluate the performance benefit of i-GELU as compared to
GELU.

% In the unusual situation where you want a paper to appear in the
% references without citing it in the main text, use \nocite
%\nocite{langley00}

\section*{Acknowledgments}
The UC Berkeley team acknowledges gracious support from Intel corporation, Intel VLAB team, Google Cloud, Google TRC team, and Nvidia, as well as valuable feedback from Prof. Dave Patterson, and Prof. Joseph Gonzalez.
Amir Gholami was supported through a gracious fund from Samsung SAIT.
Michael W. Mahoney would also like to acknowledge the UC Berkeley CLTC, ARO, NSF, and ONR.
Our conclusions do not necessarily reflect the position or the policy of our sponsors, and no official endorsement should be~inferred.

\bibliography{_main}
\bibliographystyle{icml2021}

\clearpage
\appendix

\section{Quantization Methods}
\label{appendix:quantization_methods}
\subsection{Symmetric and Asymmetric Quantization}
Symmetric and asymmetric quantization are two different methods for uniform quantization. 
Uniform quantization is a uniform mapping from floating point $x \in [x_{\mathrm{min}}, x_{\mathrm{max}}]$ to $b$-bit integer $q \in [-2^{b-1}, 2^{b-1} - 1]$.
Before the mapping, input $x$ that does not fall into the range of $[x_{\mathrm{min}}, x_{\mathrm{max}}]$  should be clipped.
In asymmetric quantization, the left and the right side of the clipping range can be different, i.e., $-x_{\mathrm{min}} \ne x_{\mathrm{max}}$.
However, this results in a bias term that needs to be considered when performing multiplication or convolution operations~\cite{jacob2018quantization}.
For this reason, we only use symmetric quantization in this work. 
In symmetric quantization, the left and the right side of the clipping range must be equal, i.e., $-x_{\mathrm{min}} = x_{\mathrm{max}} = \alpha$, 
and the mapping can be represented as~\eref{eq:uniform_quantization}.

\subsection{Static and Dynamic Quantization}
There is a subtle but important factor to consider when computing the scaling factor, $S$.
Computing this scaling factor requires determining the range of parameters/activations (i.e., $\alpha$ parameter in~\eref{eq:uniform_quantization}).
Since the model parameters are fixed during inference, their range and the corresponding scaling factor can be precomputed.
However, activations vary across different inputs, and thus their range varies. 
One way to address this issue is to use dynamic quantization, where the activation range and the scaling factor are calculated during inference.
However, computing the range of activation is costly as it requires a scan over the entire data and often results in significant overhead.
Static quantization avoids this runtime computation by precomputing a fixed range based on the statistics of activations during training, and then uses that fixed range during inference. 
As such, it does not have the runtime overhead of computing the range of activations. 
For maximum efficiency, we adopt static quantization, with all the scaling factors fixed during inference.

\section{Error Term of~\eref{eqn:lagrange}}
\label{sec:error_of_lagrange}
As one can see, the polynomial approximation of~\eref{eqn:lagrange} exactly
matches the data at the interpolating points $(x_j, f_j)$. 
The error between a target function $f(x)$ and the polynomial approximation $L(x)$ is then:
\begin{equation}
\small
|f(x) - L(x)| = \left|\frac{f^{(n+1)}(\xi)}{(n+1)!}(x - x_0) \dots (x - x_n)\right|,
\end{equation}
where $\xi$ is some number that lies in the smallest interval containing $x_0, ..., x_n$. 
In general, this error reduces for large $n$ (for a properly selected
set of interpolating points).
Therefore, a sufficiently high-order polynomial that interpolates a target function is guaranteed to be a good approximation for it. 
We refer interested readers to~\cite{stewart1996afternotes} for more details on polynomial interpolation.

\section{Experimental Details}
\label{appendix:eval_details}

\subsection{Implementation}
\label{appendix:implementation_details}
In \OURS, all the MatMul operations are performed with INT8 precision, and are accumulated to INT32 precision.
Furthermore, the Embedding layer is kept at INT8 precision. 
Moreover, the non-linear operations (i.e., GELU, Softmax, and LayerNorm) are processed with INT32 precision,
as we found that keeping them at high precision is important to
ensure no accuracy degradation after quantization.
Importantly, note that using INT32 for computing these operations has little overhead, as input data is already accumulated with INT32 precision, and these non-linear operations have
linear computational complexity.
We perform Requantization~\cite{yao2020hawqv3} operation after these operations to 
bring the precision down from INT32 back to INT8 so that the follow up operations (e.g., next MatMuls) can be performed with low precision.

\subsection{Training}
\label{appendix:training_details}

We evaluate \OURS on the GLUE benchmark~\cite{wang2018glue},
which is a set of 9 natural language understanding tasks, including sentimental analysis, entailment, and question answering.
We first train the pre-trained RoBERTa model on the different GLUE downstream tasks until the model achieves the best result on the development set.
We report this as the baseline accuracy.
We then quantize the model and perform quantization-aware fine-tuning to recover the accuracy degradation caused by quantization.
We refer the readers to~\cite{yao2020hawqv3} for more details about the quantization-aware fine-tuning method for integer-only quantization.
We search the optimal hyperparameters in a search space of learning rate $\{5\mathrm{e}-7, 1\mathrm{e}-6, 1.5\mathrm{e}-6, 2\mathrm{e}-6\}$,
self-attention layer dropout $\{0.0, 0.1\}$, and fully-connected layer dropout $\{0.1, 0.2\}$, except for the one after GELU activation that is fixed to 0.0.
We fine-tune up to 6 epochs for larger datasets (e.g., MNLI and QQP), and 12 epochs for the smaller datasets.
We report the best accuracy of the resulting quantized model on the development set as \OURS accuracy.

% MRPC~\cite{dolan2005automatically}, STS-B~\cite{cer2017semeval}, SST-2~\cite{socher2013recursive}, 
% QNLI~\cite{rajpurkar2016squad}, QQP~\cite{iyer2017first}, MNLI~\cite{williams2018broad},  
% CoLA~\cite{warstadt2018neural}, RTE~\cite{dagan2013recognizing}, WNLI~\cite{levesque2012winograd}
\subsection{Accuracy Evaluation on the GLUE Tasks}
\label{appendix:accuracy_eval}
For evaluating the results, we use the standard metrics for each task in GLUE. 
In particular, we use classification accuracy and F1 score for QQP~\cite{iyer2017first} and MRPC~\cite{dolan2005automatically},
Pearson Correlation and Spearman Correlation for STS-B~\cite{cer2017semeval},
and Mathews Correlation Coefficient for CoLA~\cite{warstadt2019neural}.
For the remaining tasks~\cite{williams2017broad, rajpurkar2016squad, socher2013recursive, dagan2005pascal}, we use classification accuracy.
For the tasks with multiple metrics, we report the average of them.
Since there are two development sets for MNLI~\cite{williams2017broad}, i.e., MNLI-match (MNLI-m) for in-domain evaluation, and MNLI-mismatch (MNLI-mm) for cross-domain evaluation, and we report the accuracy on both datasets.
We exclude WNLI~\cite{levesque2012winograd} as it has relatively small dataset and shows an unstable behaviour~\cite{dodge2020fine}.  

\subsection{Environment Setup for Latency Evaluation}
\label{appendix:env_setup}

We use TensorRT 7.2.1 to deploy and tune the latency of BERT-Base and BERT-Large models (both INT8 and FP32) on Google Cloud Platform virtual machine with a single Tesla T4 GPU, CUDA 11.1, and cuDNN 8.0. 

We should also mention that the
 most efficient way of implementing BERT with TensorRT is to use NVIDIA's plugins~\cite{tensorrtbert} that optimize and accelerate key operations in the Transformer architecture via operation fusion. 
Our estimates are that INT8 inference using NVIDIA's plugins is about 2 times faster than
na\"ively using TensorRT APIs.
However, we cannot modify those plugins to support our integer-only kernels as they are partially closed sourced and pre-compiled. 
Therefore, our latency evaluation is conducted without fully utilizing NVIDIA's plugins. 
This leaves us a chance for further optimization to achieve our latency speedup relative to FP32 even more significant. As such, one could expect the potential for a further $\sim2\times$ speed up with INT8 quantization.

\end{document}